\documentclass{article}

\usepackage{microtype}
\usepackage{graphicx}
\usepackage{subcaption}
\usepackage{booktabs} % for professional tables

\usepackage[bb=px]{mathalpha}

\usepackage{hyperref}

\usepackage[accepted]{icml2021}

\icmltitlerunning{Navigation Turing Test (NTT)}

\begin{document}

\twocolumn[
\icmltitle{Navigation Turing Test (NTT): Learning to Evaluate Human-Like Navigation}



\icmlsetsymbol{equal}{*}

\begin{icmlauthorlist}
\icmlauthor{Sam Devlin}{equal,msrc}
\icmlauthor{Raluca Georgescu}{equal,msrc}
\icmlauthor{Ida Momennejad}{equal,msrny}
\icmlauthor{Jaroslaw Rzepecki}{equal,msrc}
\icmlauthor{Evelyn Zuniga}{equal,msrc}
\icmlauthor{Gavin Costello}{nt}
\icmlauthor{Guy Leroy}{msrc}
\icmlauthor{Ali Shaw}{nt}
\icmlauthor{Katja Hofmann}{msrc}
\end{icmlauthorlist}

\icmlaffiliation{msrc}{Microsoft Research, Cambridge, UK}
\icmlaffiliation{msrny}{Microsoft Research, New York, NY, USA}
\icmlaffiliation{nt}{Ninja Theory, Cambridge, UK}

\icmlcorrespondingauthor{Sam Devlin}{sam.devlin@microsoft.com}

\icmlcorrespondingauthor{Katja Hofmann}{katja.hofmann@microsoft.com}

\icmlkeywords{Machine Learning, ICML}

\vskip 0.3in
]



\printAffiliationsAndNotice{\icmlEqualContribution} % otherwise use the standard text.

\begin{abstract}

A key challenge on the path to developing agents that learn complex human-like behavior is the need to quickly and accurately quantify human-likeness. While human assessments of such behavior can be highly accurate, speed and scalability are limited. We address these limitations through a novel automated Navigation Turing Test (ANTT) that learns to predict human judgments of human-likeness. We demonstrate the effectiveness of our automated NTT on a navigation task in a complex 3D environment. We investigate six classification models to shed light on the types of architectures best suited to this task, and validate them against data collected through a human NTT. Our best models achieve high accuracy when distinguishing true human and agent behavior. At the same time, we show that predicting finer-grained human assessment of agents' progress towards human-like behavior remains unsolved. Our work takes an important step towards agents that more effectively learn complex human-like behavior.

\end{abstract}

\vskip -.2cm
\section{Introduction}
\label{sec:intro}
\vskip -.2cm

Developing agents capable of learning complex human-like behaviors is a key goal of artificial intelligence research \cite{winston1984artificial}. Progress in this area would lead to new applications, from engaging video games \cite{soni2008bots}, to intuitive virtual assistants \cite{cowan2017can} and robots \cite{scheutz2007first}. Yet, despite striking successes in training \emph{highly skilled} agents \cite{vinyals2019grandmaster}, quantifying and making progress towards learning \emph{human-like} behavior remains an open challenge.

We propose an automated Navigation Turing Test (ANTT), a novel approach to learn to quantify the degree to which behavior is perceived as human-like by humans\footnote{All data and code to reproduce the ANTT are available at \url{https://github.com/microsoft/NTT}}. While human judges can be highly accurate when judging human-like behavior, their scalability and speed are limited. As a result, it is prohibitive to use human assessments in, e.g., iterations of training machine learning-based agents. We aim to remove this bottleneck, providing a scalable and accurate proxy of human assessments.

We demonstrate our NTT approach on the complex task of goal-directed navigation in 3D space. This task is highly relevant and practical because: (1) it has been studied from a wide range of perspectives in both biological \cite{banino2018vector,decothi2020predictive} and artificial \cite{alonso2020deep,chaplot2020object} agents, so is relatively well understood; and (2) artificial approaches achieve a high level of task success \cite{alonso2020deep}, so proficient agents can be obtained enabling us to focus on human-likeness. While the majority of research on learning 3D navigation has focused on task success, we show that a high success is not sufficient for human-like behavior.

Our key contributions compared to prior work (Sec. \ref{sec:related}) are: (1) we introduce an automated NTT for assessing human-like behavior and investigate six representative architectures that capture principal design choices of this task (Sec.~\ref{sec:method}); (2) we devise methodology for validating our automated NTT against a human Navigation Turing Test (HNTT) and empirically demonstrate its effectiveness (Sec.~\ref{sec:approach-validation}); (3) our results show high agreement of ANTT with human judges when comparing human behavior to state-of-the-art reinforcement learning agents (Sec.~\ref{sec:results}). At the same time, we demonstrate that capturing finer gradations of human-likeness in artificial agents remains an open challenge. In addition to the potential to drive learning of more human-like behavior, our work opens the door to a deeper understanding of human decision making when assessing human-like  behavior.

\vskip -.2cm
\section{Related Work}
\label{sec:related}
\vskip -.2cm

The question of how to systematically assess whether an automated agent behaves in a human-like manner was compellingly addressed by Alan Turing \cite{turing1950mind}. Initially called the ``imitation game'', and now adapted and widely known as ``Turing Test'', it asks a human judge to identify which of two respondents is human, and which is machine. Turing's initial thought experiment considered a setup where the judge could interact with each respondent through an appropriate natural language interface. Since its conception, the idea has been widely adapted to assess the degree of human-likeness in a range of scenarios, including to assess human-like play in video games \cite{yannakakis2018artificial}. Closest to our focus, a Turing Test was used to assess bot behavior in the video game Unreal Tournament as part of the 2K BotPrize Contest \cite{hingston2009turing} and human-like play in the Mario AI Championship \cite{shaker2013turing}.

While a Turing Test, if well designed, can provide strong ground truth assessments of whether humans judge behavior as human-like, several drawbacks exist. Most important from a machine learning perspective is the need to collect human judgments for every iteration of an agent, making the approach extremely costly. This makes it prohibitive for use in typical machine learning settings. Consequently, previous work has proposed automated proxies of human-likeness \cite{kirby2009companion,karpov2013believable,decothi2020predictive}.

Domain specific approaches to detecting (non) human-like behavior automatically often rely on heuristics or rules. In navigation scenarios this can include methods for detecting mistakes such as ``getting stuck'' \cite{kirby2009companion,karpov2013believable}. \citet{kirby2009companion} focus on socially acceptable robot navigation with experiments performed in simulated navigation tasks (e.g., passing people in a hallway). Human preferences were explicitly encoded as constraints (e.g., a preference for moving to the right to let someone pass), and learned behaviors were evaluated in simulation, by assessing how often a given preference was violated.
\citet{karpov2013believable} tackle the problem of ``believable'' bot navigation in the context of the BotPrize competition \cite{hingston2009turing}. Believable behavior is assessed using rules for detecting mistakes such as getting stuck in cyclic behavior, collisions with other entities, failing to move, or falling off a map. A key drawback of rule-based automation is that they require high manual effort by domain experts to construct. In addition, due to this high effort, they are necessarily coarse and unable to capture nuanced aspects of human-likeness.

Addressing both the limitations of scalability of human Turing Tests and rule-based approaches, learned models can serve as a proxy measure of human-likeness. If such a model can learn accurate predictions of human-likeness given ground truth judgments, these approaches could be deployed rapidly and at scale to drive the development and training of more human-like agents. Closely aligned with this goal of our work is the recent development of imitation learning approaches that rely on a discriminator to learn to distinguish between learned imitation and demonstrated expert behavior \cite{ho2016generative,peng2018variational}. However, while closely aligned with the goals of our work in principle, the current state of the art in this field has not yet reached the level required for empirically strong performance on complex tasks and human-like behavior, with many evaluated on relatively less complex behaviors and imitating other automated agents. Our work has the potential to pave the way towards learning complex human-like behavior, and provides important insights into the key design choices required for capturing notions of human-likeness. We include representative examples closest to this approach as baselines (model SYM-FF and VIS-FF, Tbl.~\ref{table:models-overview}).

Closely related to our work is the approach taken by \citet{deCothi2020.09.26.314815} to compare artificial agent behavior in a maze navigation task to human and rat behavior. They used a support vector machine classifier trained on a small set of expert designed features to distinguish between model-free, model-based, and successor-representation-based trained agents. The trained classifier was then applied to behavior of humans and rats to estimate which agent behavior was closest to that of biological agents \cite{deCothi2020.09.26.314815,decothi2020predictive}. We base our approach on this setup, but extend it to take as input lower-level features (e.g., trajectories, images) to avoid the need for expert designed feature extraction and better align with the line of work on imitation learning that forms the second inspiration to our work.

While our approach makes no specific assumption about the type of task for which we aim to detect human-like behavior, we focus on the task of navigation in 3D space as it has been widely studied from a range of perspectives, including neuroscience \cite{iaria2003cognitive,banino2018vector,decothi2020predictive} and machine learning \cite{alonso2020deep,chaplot2020object}. In addition, recent years have seen rapid progress on machine learned agents that achieve high task success, allowing us to focus our attention on whether the learned behavior is indeed human-like. A driver of progress on learning to navigate has been the availability of simulated environments and evaluation frameworks. \citet{anderson2018evaluation} survey recent progress and evaluation frameworks for 3D navigation. In line with the majority of work in this area, the focus is on evaluating task success. Recent work on learning to navigate with a machine learning focus includes improvements of sample efficiency, e.g., through unsupervised pretraining \cite{li2020unsupervised} and object semantics \cite{chaplot2020object}. Here we leverage the work of \citet{alonso2020deep}, as it most closely aligns with our 3D game environment. At the intersection of machine learning and neuroscience, machine learned models have been proposed as models for studying biological behavior \cite{banino2018vector,decothi2020predictive}. As before, the primary evaluation of these focused on task success \cite{banino2018vector}, with the notable exception of \citet{deCothi2020.09.26.314815} which forms the starting point of our work.

\vskip -.2cm
\section{Method}
\label{sec:method}
\vskip -.2cm

In this work we approach the task of predicting human judgments of whether an observed behavior appears human-like as a classification task, as detailed in Section \ref{subsec:problem}. Our key focus is the question of which type of input space (Sec.~\ref{subsec:input}) and model (Sec.~\ref{subsec:models}) can best capture human judgments.\footnote{See Appendix A.1 for training details and hyperparameters.}

\vskip -.2cm
\subsection{Problem Setup}
\label{subsec:problem}
\vskip -.2cm

We formalize the task of assessing human-likeness as binary classification. We assume input data in the form of time-indexed trajectories $x_{0:T}$ of length $T$, where $x_t \in \mathbb{R}^{d}$ denotes the $d$-dimensional observation at time step $t$.

Our goal is to obtain a classifier that achieves high alignment with ground truth human judgments (detailed in Sec.~\ref{subsec:ground-truth}). Ground truth takes the form $y^* : x^{(1)} \succ x^{(2)}$ to indicate that a single human judged the trajectory represented by $x^{(1)}$ to be more human-like than $x^{(2)}$. We obtain repeat measurements from several human judges for each pair. We focus on two evaluation measures: (1) accuracy compared to human ground truth (aggregated by majority vote), and (2) Spearman correlation (a non-parametric rank correlation measure robust to outliers \cite{croux2010influence}) compared to a ranking of agent trajectories by the proportion of judges that assessed them as human-like. Together, these capture coarse and fine-grained agreement between our classification models and human judgments of human-likeness.

In this work, we assume that human ground truth judgments are only available at test time. This is a realistic assumption given the resource intensive nature of this type of data. To address this challenge, we propose training our models with  \emph{proxy labels}, which are derived from known agent identity $\hat{y}$. Thus, our classifiers are trained to maximize accuracy when predicting whether a trajectory was truly generated by a human player or an artificial agent, using a binary cross entropy loss:
$L = -[\hat{y}_{H=1} \log(p_{H=1}) + (1 - \hat{y}_{H=1}) \log(1-p_{H=1})]$,
where we denote by $\hat{y}_{H=1}$ the pseudo label indicating that trajectory was truly generated by a human player, and by $p_{H=1}$ the probability that the current model assigns to the trajectory having been generated by a human player. 

A key question is whether the resulting models can generalize to align well with the degree to which human judges actually \emph{perceive} a trajectory as more or less human. A key design choice that we expect to influence this ability to capture human judgments is the input space the classifier operates on. We discuss these choices in the next section.

\vskip -.2cm
\subsection{Choice of Input Space}
\label{subsec:input}
\vskip -.2cm

Classifier input can be represented in a variety of ways. For example, presenting input in the form of video frames derived directly from the game footage, is likely to align well with the information available to human judges when they decide on whether a given behavior appears human-like. We hypothesize that observation spaces that are well aligned with the information available to human judges will result in higher accuracy on the test task of capturing human judgments of human-like behavior. However, more complex observations, such as video, may suffer from low amounts of training data, leading to a potential trade-off between model fidelity and sample efficiency.

\begin{figure}[t]
\vskip -.3cm
\begin{minipage}[c][7.5cm][t]{.49\columnwidth}
  \vspace*{\fill}
  \centering
  \includegraphics[width=3.2cm,height=7.3cm]{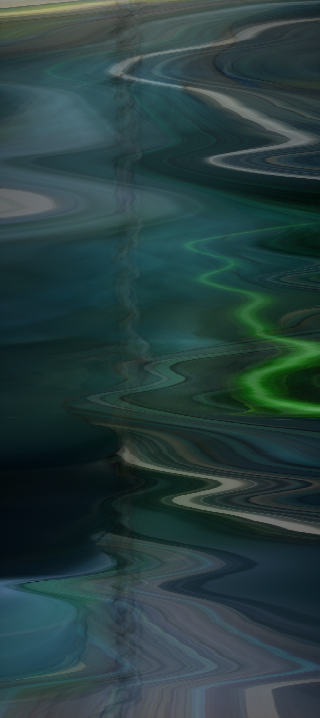}
  \subcaption{Bar-code input.}
  \label{fig:barcodes-trajectory}
\end{minipage}%
\begin{minipage}[c][8.3cm][t]{.49\columnwidth}
  \vspace*{\fill}
  \centering
  \includegraphics[width=3.2cm,height=2.8cm]{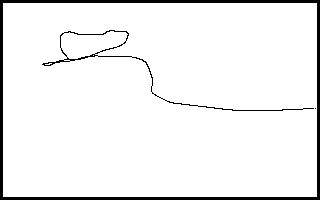}
  \subcaption{Sample of a top-down trajectory ($320\times200$ pixels).}
  \label{fig:top-down-trajectory}\par\vfill
  \includegraphics[width=3.8cm,height=3.2cm]{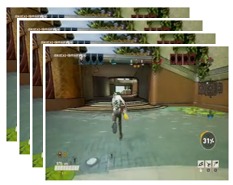}
  \subcaption{Video frames ($320\times200$)}
  \label{fig:video-trajectory}
\end{minipage}
\vskip -.2cm
\caption{Examples of the inputs used by our classifiers: (\subref{fig:barcodes-trajectory}) bar-code of a video (used by BC-CNN), (\subref{fig:top-down-trajectory}) top-down trajectory (TD-CNN), (\subref{fig:video-trajectory}) video (frame sequence, VIS-FF and VIS-GRU).}
\label{fig:inputs-trajectories}
\vskip -.2cm
\end{figure}

We consider the following choices (samples in Fig.~\ref{fig:inputs-trajectories}):

\textbf{SYM.} \emph{Symbolic} observations provide a low-dimensional representation of each step along the agent trajectory. Here, we use absolute agent location in 3D space as it provides a minimal representation of agent state.

\textbf{VIS.} \emph{Visual} observations provide images of size 320x200 of the actual game footage, rendered without overlays such as game score. This observation type is expected to be well aligned with the information available to a human judge when assessing whether a behavior appears human-like.

In our architectures, we explore feed-forward and recurrent models for encoding observations, as detailed in Section \ref{subsec:models}. In addition, we consider two compressed representations that abstract the respective observations in a single image:

\textbf{TD.} \emph{Top-down} trajectory projection (a single 2D image per trajectory), obtained by projecting the \emph{symbolic} representation (agent position) along the "up" direction (z-coordinate).

\begin{table*}[t]
\centering
\vskip -.2cm
\resizebox{1.98\columnwidth}{!}{
\begin{tabular}{llp{.4\columnwidth}lp{.9\columnwidth}}
\toprule
Name & Input (Sec.~\ref{subsec:input}) & Encoder & Pre-Training &  Notes\\
\midrule
SYM-FF & \emph{Symbolic} & Feedforward Network & [None] & Equivalent to typical discriminator architectures in adversarial imitation learning \cite{ho2016generative} \\
SYM-GRU & \emph{Symbolic} & GRU \cite{cho2014properties} & [None] & \\
\midrule
VIS-FF & \emph{Visual} & VGG-16 \cite{simonyan2014very} & Imagenet (CNN) & Equivalent to discriminators for adversarial imitation learning from visual observations \cite{li2017inferring}\\
VIS-GRU & \emph{Visual} & CNN (VGG-16) + GRU & Imagenet (CNN) &  \\
\midrule
TD-CNN & \emph{Top-down} & CNN (VGG-16) & Imagenet (CNN) & \\
BC-CNN & \emph{Bar-code} & CNN (VGG-16) & Imagenet (CNN) & \\
\bottomrule
\end{tabular}
}% end resizebox
\vskip -.3cm
\caption{Overview of the ANTT model architectures investigated in this work. See Sections \ref{subsec:input} and \ref{subsec:models} for details.}
\vskip -.2cm
\label{table:models-overview}
\end{table*}

\textbf{BC.} \emph{Bar-code} encodes \emph{visual} observations (a single 2D image per trajectory video). It is obtained by collapsing every video frame into a single line by averaging along the y-dimension of the frame, and then stacking those lines into a single image. If the input video had 600 frames of size 320x200, the resulting bar-code has size 320x600.

We hypothesize that the compressed TD and BC representations capture key patterns of human-likeness while providing higher information density than symbolic or visual observations, leading to more effective classification.

\vskip -.2cm
\subsection{Models}
\label{subsec:models}
\vskip -.2cm

Following from our problem definition above (Section \ref{subsec:problem}), we now detail our proposed classification models. Each model tests specific hypotheses about how to best learn to align with human judgments. A summary of all models is provided in Table \ref{table:models-overview}, with details as follows:

\textbf{FF.} Feedforward models extract features from individual states, without taking into account the sequential structure along a trajectory. We construct this type of model as a representative baseline in line with discriminators typical of adversarial imitation learning \cite{ho2016generative}.

\textbf{GRU.} Recurrent models explicitly model sequence information and are expected to more directly align with human assessments (human assessors view observations in the form of videos, c.f., Section \ref{sec:approach-validation}). We encode sequences using Gated Recurrent Unit (GRU) networks \cite{cho2014properties}.

\textbf{CNN.} Convolutional models are applied to image input (visual, top-down and bar-code observations). We use a VGG-16 \cite{simonyan2014very} pre-trained on Imagenet \cite{deng2009imagenet} to extract visual features.

For all model training, input batches of individual observations (feedforward) and sequences (GRU) are sampled uniformly from trajectories. At test time, predictions are aggregated per trajectory using the majority vote. Learning aggregation at the trajectory level is an important direction of research, however improving simple and robust aggregation baselines with learned approaches is challenging \cite{radford2018sequence,lan2020learning}.

\begin{figure}[th]
%\vskip 0.2in
\begin{center}
\centerline{\includegraphics[trim=60 0 
0 50,clip,width=.96\columnwidth]{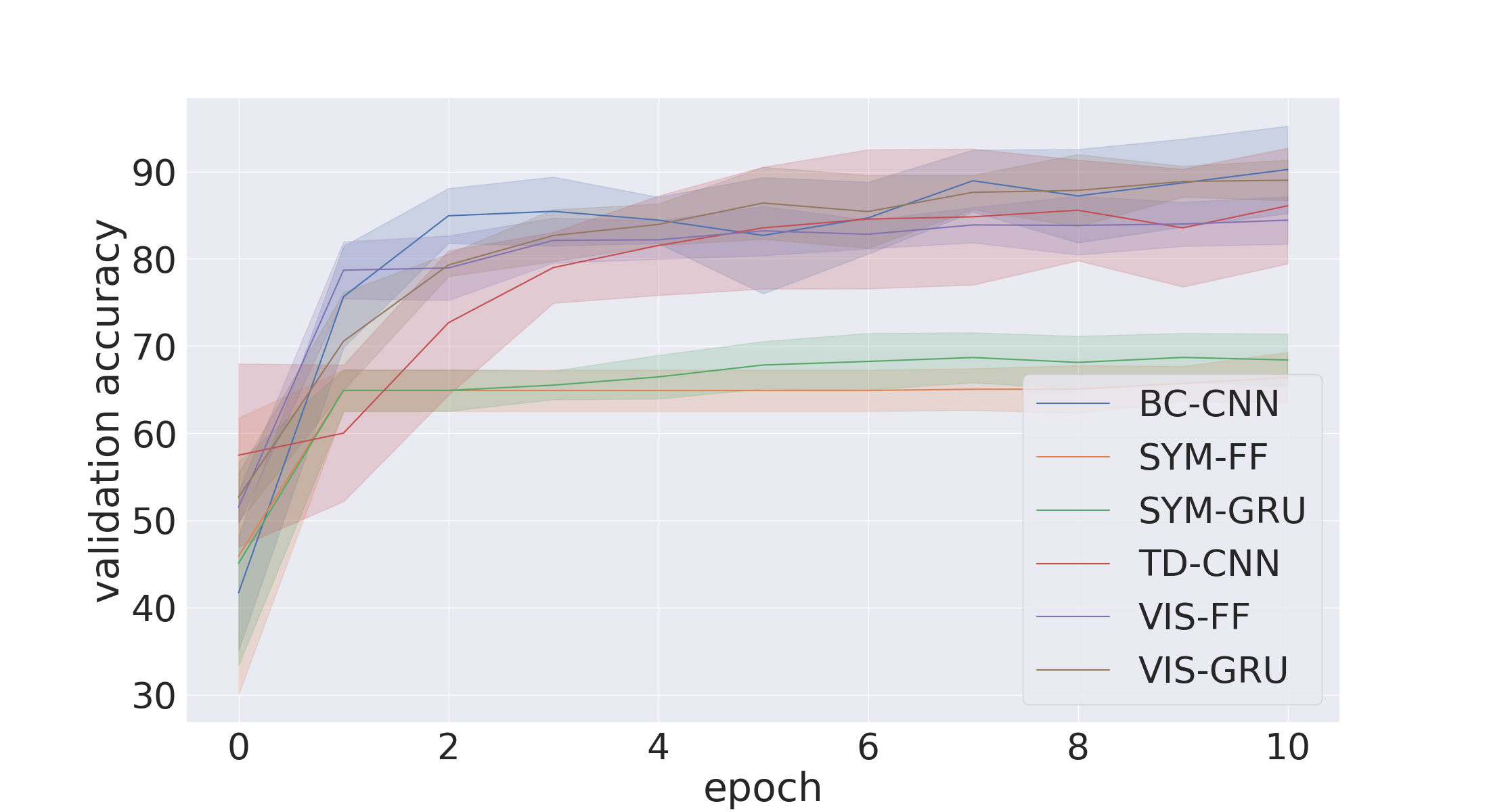}}
\vskip -.4cm
\caption{Validation accuracy over training epochs for our ANTT models (5-fold cross validation). Symbolic models achieve relatively lower validation accuracy, however, our later results show they are less prone to overfitting (See Section.~\ref{subsec:res-main}).
}
\label{fig:classifier-validation}
\vskip -.9cm
\end{center}
\end{figure}

Before moving forward, we validate that the proposed models are effective on the task they are trained to perform - distinguishing between behavior generated by human vs artificial agents. The validation curves in Figure \ref{fig:classifier-validation} show that all proposed models learn to distinguish between these on our validation set. This confirms that we have a strong basis for evaluating which of the proposed approaches matches ground truth provided by human assessors. Our main results on this question will be presented in Section \ref{sec:results}, after we detail our approach to validating the classifiers in Section \ref{sec:approach-validation}. 

\vskip -.2cm
\section{Validating ANTT by Human NTT}
\label{sec:approach-validation}
\vskip -.2cm

This section details our approach to validating the methods proposed for automatically evaluating human-like behavior (Sec.~\ref{sec:method}) with a human NTT (HNTT). The following components are required for this validation, and are discussed in turn below. First, we detail the 3D navigation task on which human-like behavior is assessed (Sec.~\ref{subsec:task}). Second, we need to collect behavior data from human players (Sec.~\ref{subsec:human-nav}) and artificial agents (Sec.~\ref{subsec:agent-nav}). Third, we obtain ground truth human judgments of human-likeness (Sec.~\ref{subsec:ground-truth}).\footnote{See Appendices for further details: A.2 (agent training), A.3 (human data collection) and A.4 (evaluation).}

\vskip -.2cm
\subsection{Definition of the 3D Navigation Task}
\label{subsec:task}
\vskip -.2cm

\begin{figure}[t]
%\vskip 0.2in
\begin{center}
\centerline{\includegraphics[width=.95\columnwidth]{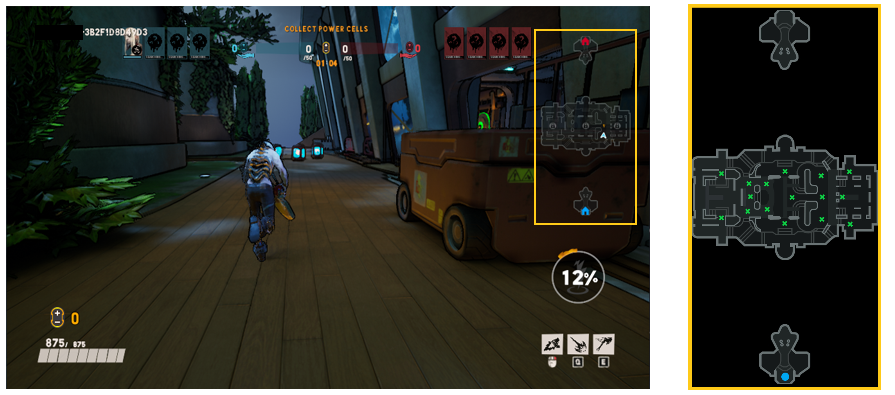}}
\vskip -0.1in
\caption{Screenshot of the 3D navigation task as perceived by human players (left). Mini map of the level (right), indicating as a blue circle the player's spawning location (bottom) and in green the 16 possible goal locations. The main map is approximately 500x320 meters in size and includes complex 3D elements.  The screenshot is not representative of the final game play and visuals.}
\label{fig:be-screenshot}
\end{center}
\vskip -0.4in
\end{figure}

Our experiments are conducted in a modern AAA video game.\footnote{We thank our colleagues at Ninja Theory for providing access to this game for research purposes only.} 
A screenshot of the navigation task is shown in Figure \ref{fig:be-screenshot}, together with a top-down view of the map. The task is set up on a map that is representative in size and complexity for a modern 3D video game, and comparable to recent navigation experiments, such as the large game map explored by \citet{alonso2020deep}. The players' spawning position is on an island outside the main map, connected by a set of 3 available jump areas. Agents must learn to navigate to a target location on the main map, represented by the blue containers in the left screenshot in Figure \ref{fig:be-screenshot}. On each navigation episode, the target spawns uniformly at random in one of 16 possible locations. In preliminary tests we confirmed that human players can successfully navigate to all goal locations within $26$ seconds on average.

\vskip -.2cm
\subsection{Collecting Human Navigation Data}
\label{subsec:human-nav}
\vskip -.2cm

We now describe how we collected human navigation data.

\textbf{Requirements.} Data from a total of 7 human players was collected using the replay capture engine in an experimental game build implementing the navigation task (Section \ref{subsec:task}). Human players' expertise with video games varied, ranging from casual to advanced players. All players had high familiarity with the game map and task, and recorded replays on machines that met the system requirements of the experimental game build, including GPU rendering support.

\textbf{Parameters and Constraints.} The navigation target was indicated on-screen in a mini-map that was shown throughout the navigation task (Fig.~\ref{fig:be-screenshot}). There was a limited action space: agents and human players could only navigate (e.g., jump, attack and special abilities were disabled). The human players used keyboard and mouse navigation only. 
    
\textbf{Video Generation.} The episode replays were saved as MP4. A total of 140 human recordings were collected, which we split into 100 videos (from 4 players) for classifier training and validation, and 40 (3 remaining players) for testing.

\textbf{Post-Processing.} We applied post-processing edits to the videos for final inclusion into the HNTT ground truth assessment (Section \ref{subsec:ground-truth}). These edits are defined by a set of rules which were uniformly applied to all the videos. \textbf{Rule 1:} Mask identifying information (computer name tag). \textbf{Rule 2:} Cut all videos to start the moment the character lands on the main map.\footnote{This is to focus our current validation on ground-based navigation. Assessing human-like flying is left for future work.} 

\textbf{Rule 3:} Cut the last 1-3 seconds of the human player videos to just before they stopped moving to correct for an effect of the human data collection process, where human players had to manually end the recording and thus appeared briefly ``stopped" at the end of each episode.

\vskip -.2cm
\subsection{Collecting Agent Navigation Data}
\label{subsec:agent-nav}
\vskip -.2cm

This section details training of the artificial agents used to generate agent navigation data, including agent architectures, training process, and agent navigation data collection.

\begin{figure}
\begin{minipage}[c][7cm][t]{\columnwidth}
  \centering
  \includegraphics[width=8.2cm,height=2.4cm]{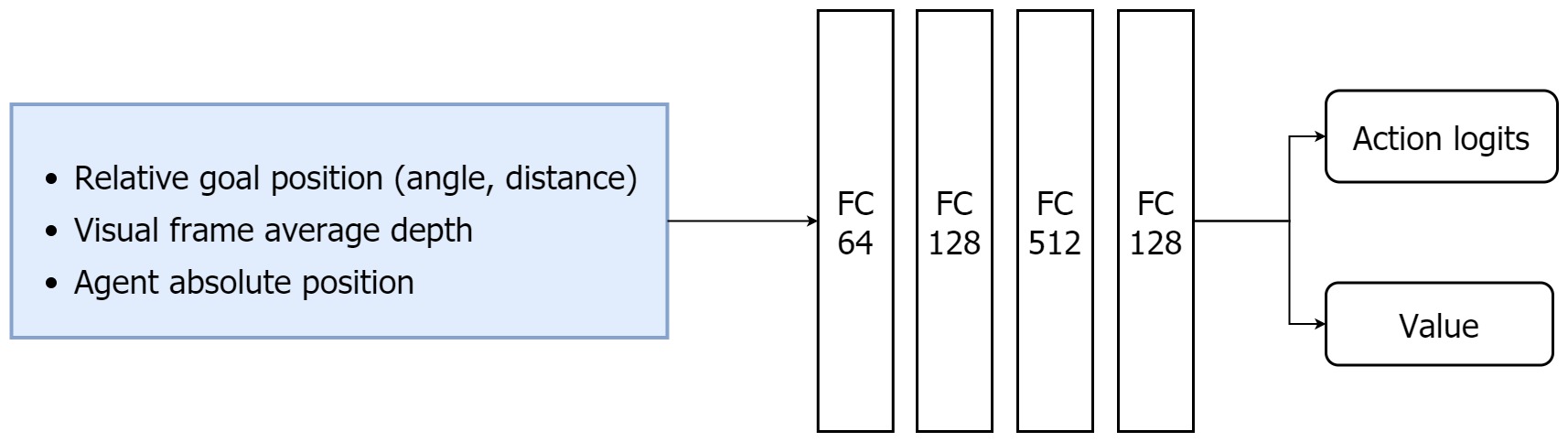}
  \subcaption{Symbolic agent model architecture}
  \label{fig:symb-model}\par\vfill
  \includegraphics[width=8.2cm,height=3.2cm]{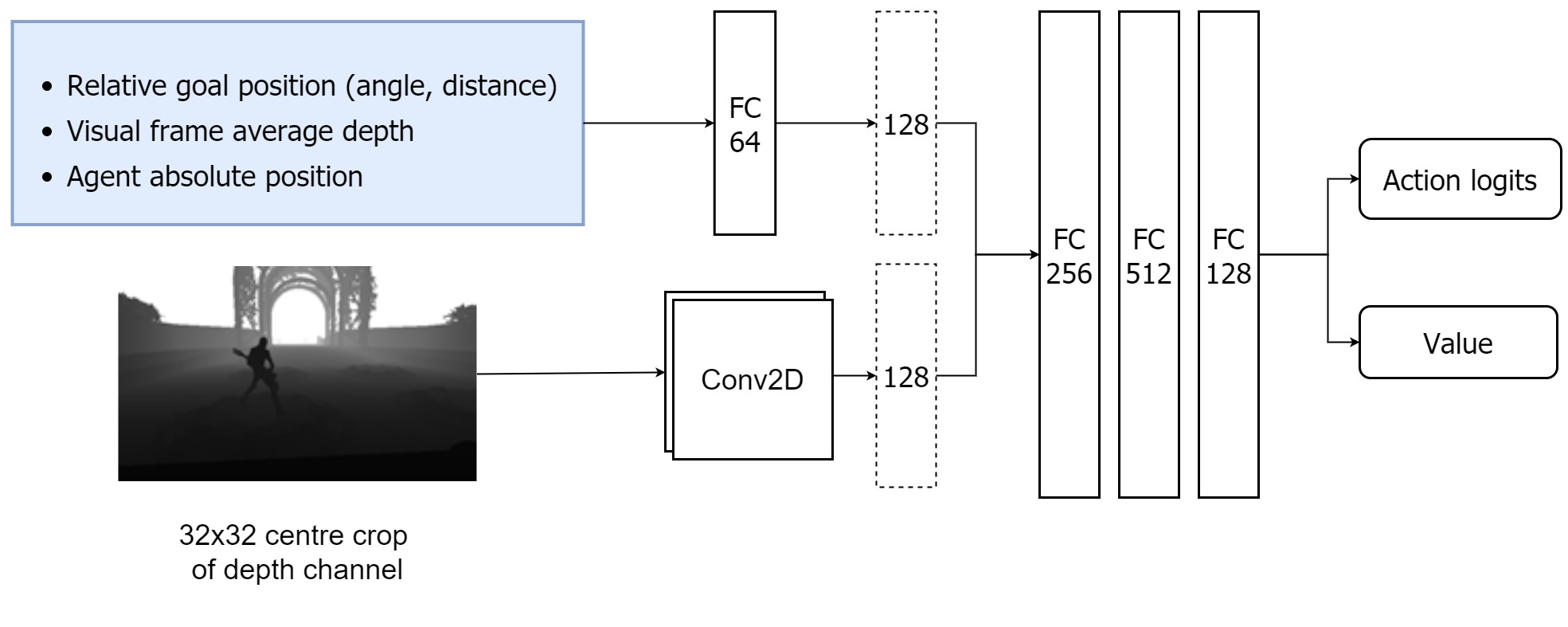}
  \subcaption{Hybrid agent model architecture}
  \label{fig:hybird-model}
\end{minipage}
\vskip -.2cm
\caption{Overview of the agent architectures used in this study.}
\vskip -.2cm
\label{fig:be-models}
\end{figure}

Figure \ref{fig:be-models} shows our two reinforcement learning agent architectures: symbolic and hybrid. Both are representative examples of current state-of-the-art approaches to learning 3D navigation in video games, closely following very recent work by \citet{alonso2020deep}. We chose this starting point as a representative approach that learns effective navigation in a game similar to ours, and select two observation spaces and architectures for which we expect high task performance based on the ablation studies of \citet{alonso2020deep}.

\textbf{Symbolic.}
The \emph{symbolic} agent (Fig.~\ref{fig:symb-model}) observes a low-dimensional representation of the goal (relative angle and distance) and game state (agent position, average frame depth), processed using a feed forward network. We hypothesize that this compact observation space enables sample-efficient learning but relatively less human-like behavior.

\textbf{Hybrid.}
The \emph{hybrid} agent (Fig.~\ref{fig:hybird-model}) receives the symbolic observation, as above, and a 32x32 depth buffer of the game footage. Image features are extracted using a CNN before concatenating with the output of the feed forward layers that process symbolic features. Given access to (coarse) visual and depth information, we hypothesize the agent to be better aligned with the type of information used by human players and to consequently learn more human-like behavior.

Both agents use a discrete action space: moving forwards, left and right to a set of given degrees (30, 45, 90), as well as turning in place. Both agents receive the same reward signal during training. This consists of a dense reward for minimizing the distance between the goal location and the agent's position, a +1 reward for reaching the target, and a -1 penalty for dying, as agents are able to jump off the edges of the map. In addition, a small per-step penalty of $-0.01$ encourages efficient task completion. Episodes end when agents reach the goal radius or after $3,000$ game ticks ($\approx 50$ seconds, $210$ agent steps).

The agents were trained using PPO \cite{schulman2017proximal}, implemented in Tensorflow \cite{tensorflow2015-whitepaper}. Behavior policy rollouts during training were collected through a scalable distributed framework on $60$ parallel game instances. 
In each training episode, the agent spawned in one of the 26 predefined locations, both on the main map and in the spawn area, with a $34\%$ preference for the latter. This uniform domain randomization \cite{tobin2017domain} was found sufficient for sample efficient learning in preliminary experiments.

\begin{figure}[t]
\vskip -0.1in
\begin{center}
\centerline{\includegraphics[trim=80 0 120 30,clip,width=.99\columnwidth]{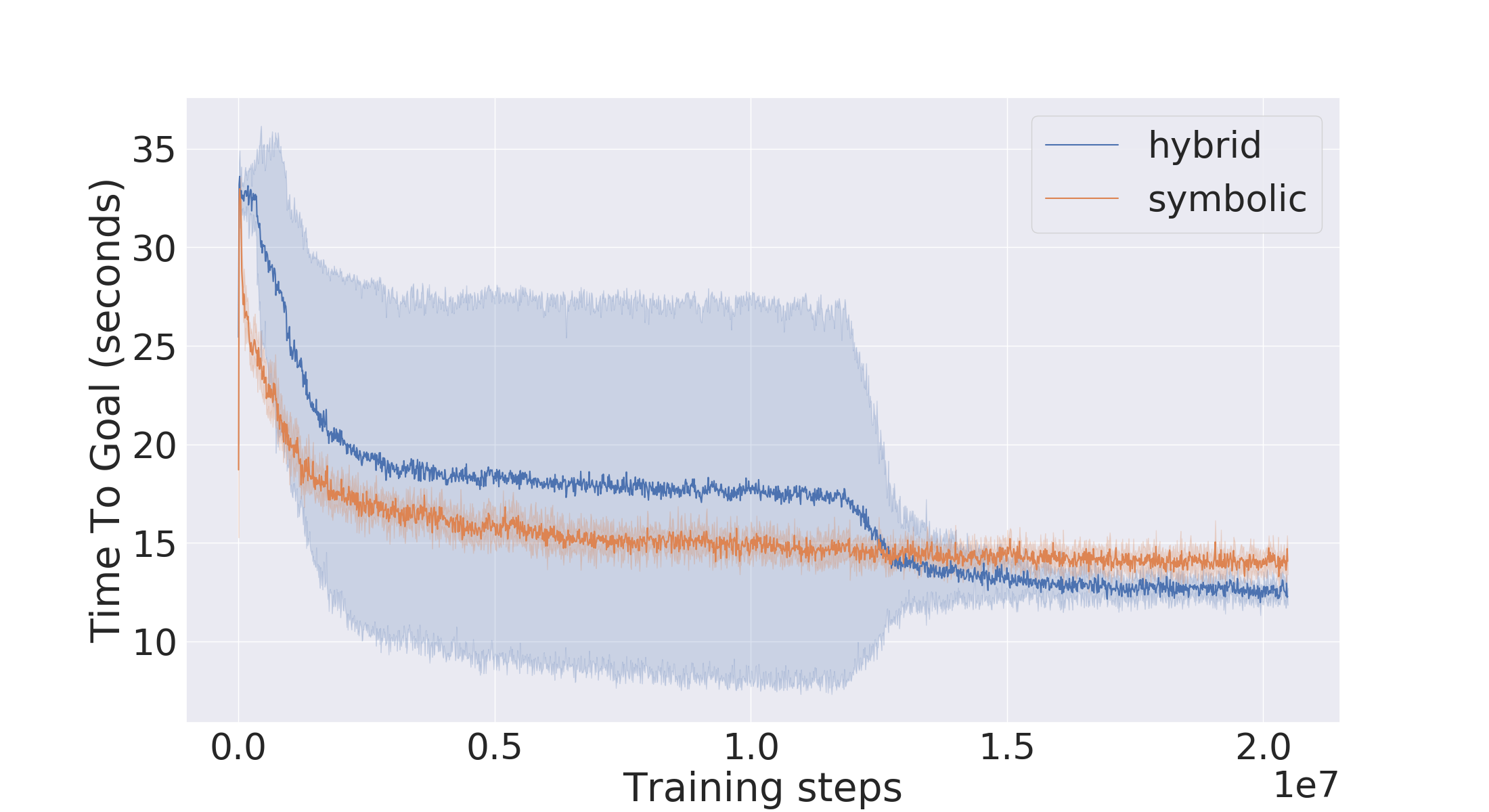}}
\vskip -0.1in
\caption{Time to goal over training steps (5 seeds each). Both models achieve high task success. The high variance observed in the hybrid agent is due to one seed that had a delay in convergence.}
\label{fig:agent-success}
\end{center}
\vskip -0.3in
\end{figure}

Before proceeding to the next stage, we verify that the selected agent models successfully learn the navigation task. Figure \ref{fig:agent-success} verifies that the agent models used here are representative of the state of the art of learning to navigate in complex 3D games. As shown, agents learn to reliably reach the goal within the first 5M training steps. On average, the hybrid agent achieves task completion 10\% faster than the symbolic agent. This demonstrates that our trained agents achieve the required level of task success, allowing us to focus on whether the learned behavior is also human-like.

To capture agent navigation data we rolled out the model checkpoints after 20M training steps. The rolled out policies are non-deterministic. The starting position for all rollouts was the spawn area (as for human players). The episodes were saved as videos and then sampled to match the navigation goals in the selected human navigation samples. The agent navigation was post-processed in line with the human navigation data (Sec.~\ref{subsec:human-nav}).

\vskip -.2cm
\subsection{Collecting Ground Truth Human Judgments}
\label{subsec:ground-truth}
\vskip -.2cm

We designed two behavioral human NTT (HNTT) studies to establish ground truth based on collecting human judgments of agent vs. human navigation in the game. Each study was administered in the form of an online survey. 

\textbf{IRB Approval.} The survey was approved by our institution's Institutional Review Board (IRB), authorizing us to collect data from internal employees. An IRB approved digital consent form was obtained before participation.

\textbf{Participants.} A total of 60 human assessors participated in the HNTT. We recruited participants from internal mailing lists, none of the participants were familiar with our study. Participants were required to be 18 years or older. No additional demographic information was collected. Participants were asked to indicate their familiarity with third-person action games and the game title used in this study. 

\begin{figure}[t]
%\vskip 0.2in
\begin{center}
\fbox{
\begin{minipage}{7.5 cm}
\centerline{\includegraphics[width=\columnwidth]{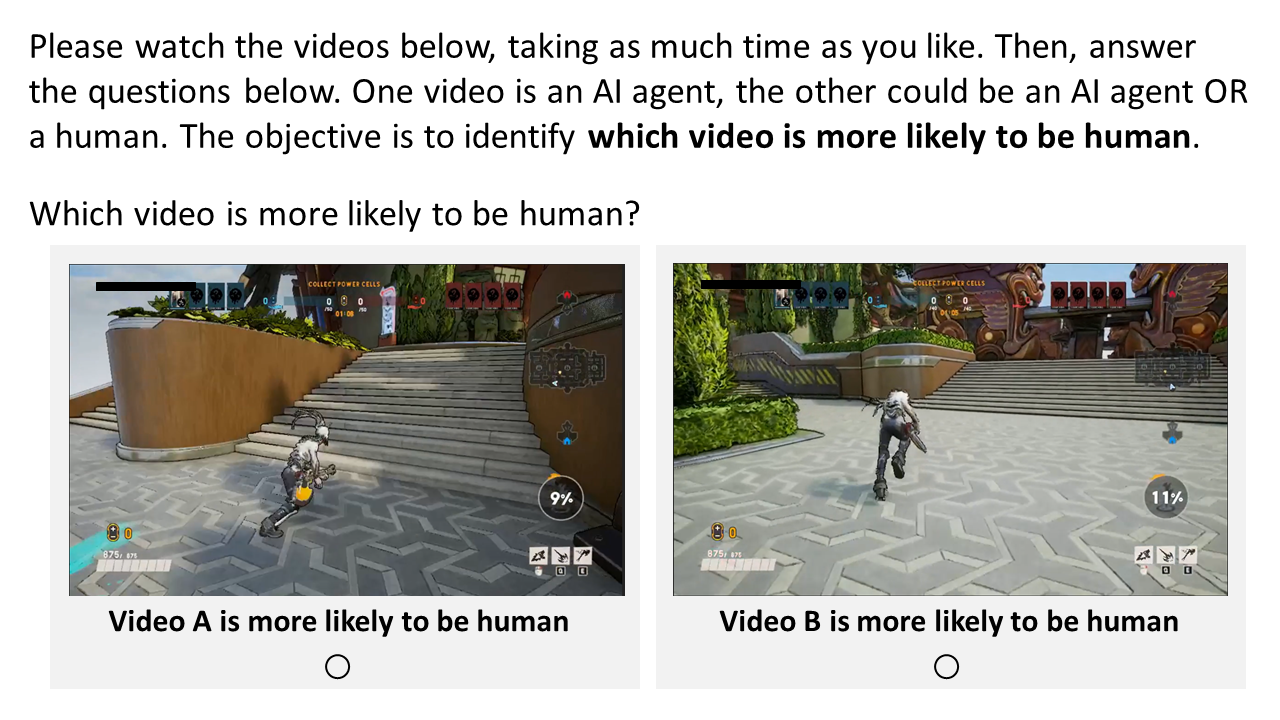}}
\centerline{\includegraphics[width=\columnwidth]{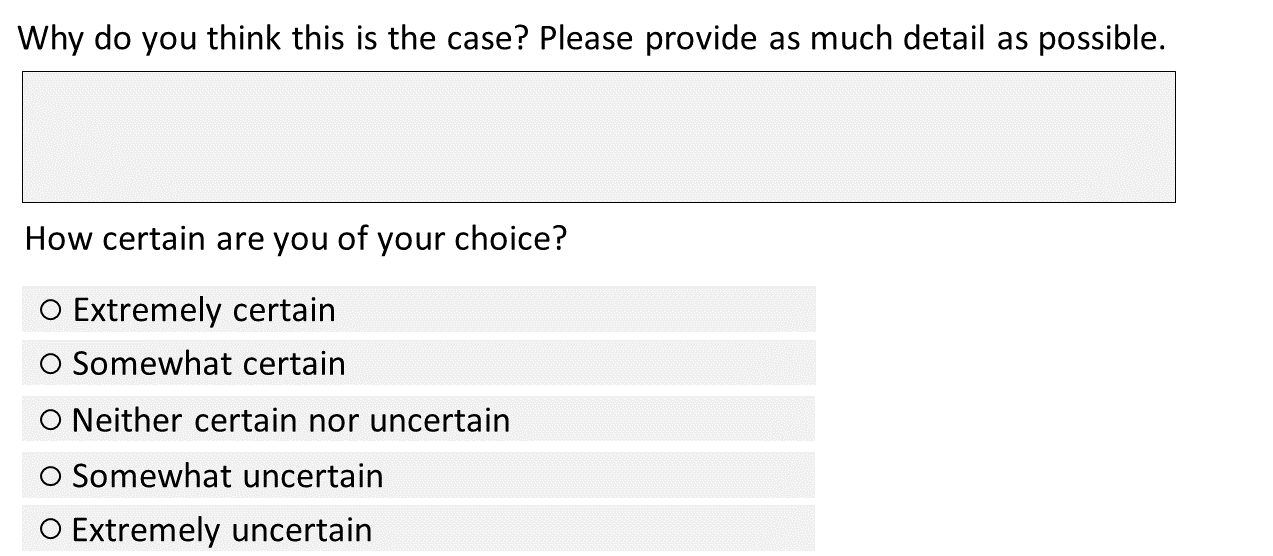}}
\end{minipage}
}
\caption{Example of a Human Navigation Turing Test (HNTT) trial. Videos are not representative of actual game play and visuals.}
\label{fig:study-interface}
\end{center}
\vskip -0.2in
\end{figure}

\textbf{Study Design.} Each HNTT study consisted of 10 Turing Test trials, in each of which the stimuli consisted of two side-by-side videos, A and B, of an avatar navigating a game environment (Figure \ref{fig:study-interface}). The task of the participant was to watch each video and answer the question ``Which video is more likely to be human?", explain their rationale, and indicate uncertainty of their decision on a 5-point Likert scale varying from ``extremely certain" to ``extremely uncertain".

In each study all participants viewed the same ten HNTT trials with the same stimuli, presented in randomized order per participant (within-subject design). Videos A and B were selected from a set of short videos in which the symbolic or hybrid agent (Section \ref{subsec:agent-nav}), or a human player (Section \ref{subsec:human-nav}) completed the navigation task (Section \ref{subsec:task}).

\textbf{Stimuli and conditions.} The presented  video stimuli and trial conditions differed between Study 1 and Study 2 according to Table \ref{table:2}, administered as follows.
   
%\begin{center}
\begin{table}[h!]
\centering
\resizebox{.75\columnwidth}{!}{
\begin{tabular}{ c | c  c  c }
\toprule
 vs.  & Human & Hybrid & Symbolic\\ 
\midrule
Human & - & Study 1  & Study 2  \\ 
\midrule
Hybrid agent & Study 1 & - & Study 1,2    \\
\midrule
Symbolic agent & Study 2 & Study 1,2 & - \\
\bottomrule
\end{tabular}
}% end resizebox
\caption{Table of HNTT comparison conditions in Study 1 and Study 2. Each study consisted of 6 human vs. agent comparison trials, and 4 trials comparing the two agents.}
\label{table:2}
\end{table}

   \textbf{Study 1.} Human participants were asked to judge a human player vs. a hybrid agent in 6 trials with side-by-side video pairs, and a symbolic agent vs. a hybrid agent in 4 trials with side-by-side video pairs (Table \ref{table:2}).
   
   \textbf{Study 2.} Human participants were asked to judge a human player vs. a symbolic agent in 6 trials with side-by-side video pairs, and a symbolic agent vs. a hybrid agent in 4 trials with side-by-side video pairs (Table \ref{table:2}).
   
The video stimuli used in the study were sampled uniformly at random from our available recordings, matching the video pairs on task goal location. See Figure \ref{fig:study-interface} for a screenshot. % of a trial.

\vskip -.2cm
\section{Results}
\label{sec:results}
\vskip -.2cm

In this section we detail our main empirical results of two Navigation Turing Test studies, obtained according to the validation approach detailed above (Sec.~\ref{sec:approach-validation}).

\vskip -.2cm
\subsection{HNTT: Human Assessments of Human-Likeness}
\label{subsec:res-groundtruth}
\vskip -.2cm

\begin{figure}[t]
\vskip -.06in
\begin{center}
\includegraphics[trim=0 0 0 0,clip,width=.5\columnwidth]{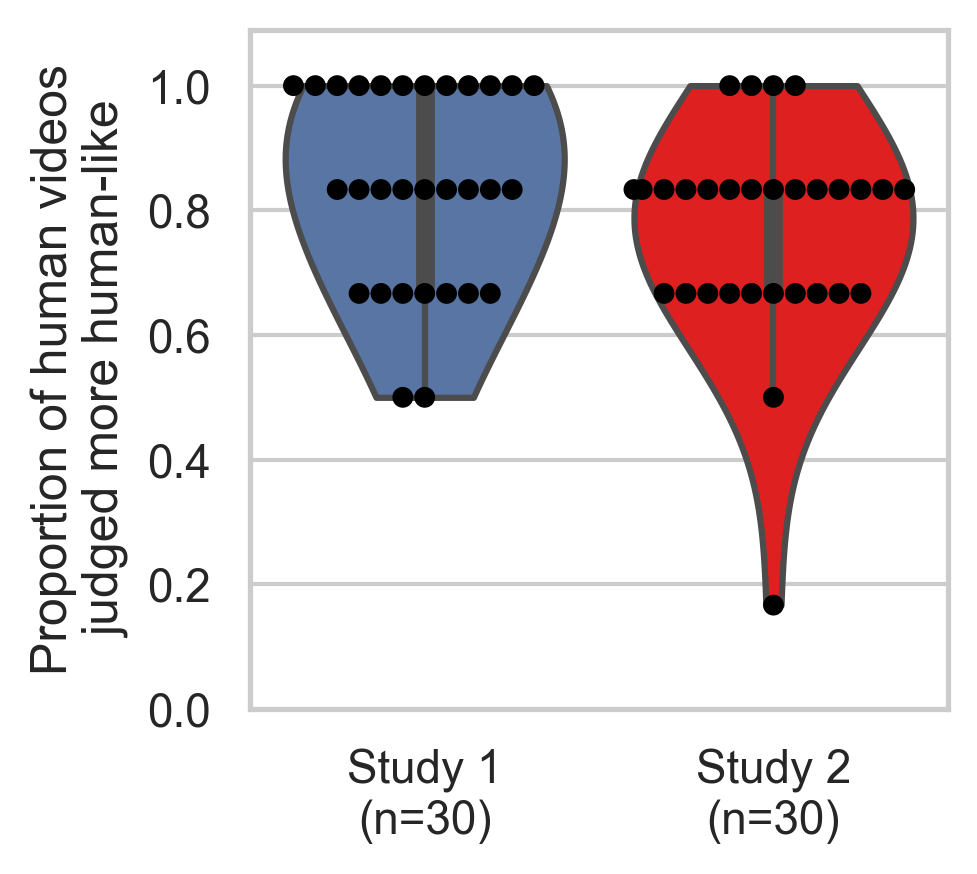}
\includegraphics[trim=0 0 0 18,clip,width=.46\columnwidth]{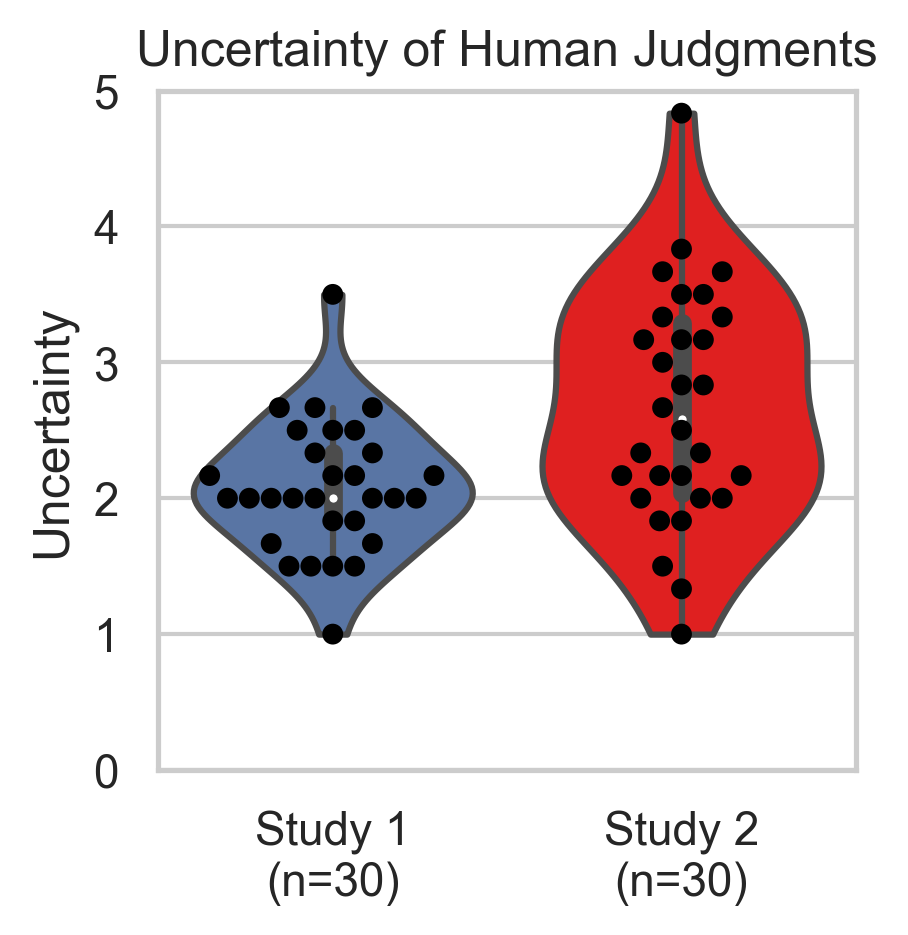}
\vskip -0.15in
\caption{HNTT Results. Left: violin plots display proportion of videos with true human behavior (ground truth) that were judged more human-like. Each participant's agreement with ground truth is a dot in the swarm plot. Right: Reported uncertainty for the same judgment (1: extremely certain to 5: extremely uncertain).}
\label{fig:ground-truth-human-agent}
\end{center}
\vskip -0.1in
\end{figure}

\begin{figure}[t]
\vskip -.06in
\begin{center}
\includegraphics[trim=5 0 115 20,clip,width=.52\columnwidth]{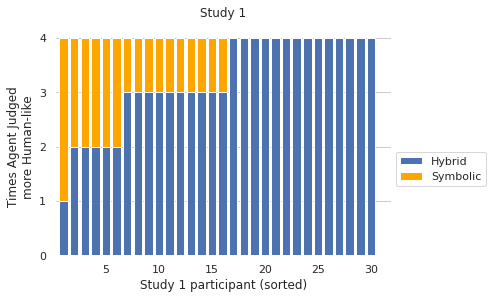}
\includegraphics[trim=50 0 10 20,clip,width=.47\columnwidth]{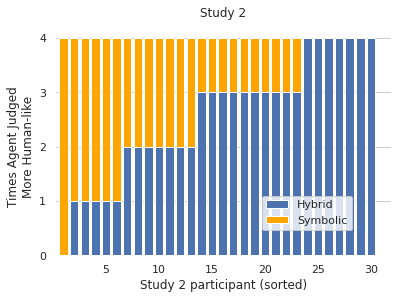}
\vskip -0.2in
\caption{Responses in direct comparisons of symbolic and hybrid agents in Study 1 (left) and 2 (right). Y-axis: number of times a participant judged an agent more human-like (out of $4$). X-axis: participants, ordered by how often they chose the hybrid agent.}
\label{fig:ground-truth-hybrid-symbolic}
\end{center}
\vskip -0.2in
\end{figure}

We analyzed human behavioral data from two human NTT (HNTT) studies, in each of which participants observed 10 pairs of video stimuli, judged which is more human-like, and marked their uncertainty. These results were used to establish ground truth for automated NTT studies (Sec.~\ref{subsec:ground-truth}). We first assessed the extent to which human participants accurately judged videos of real human players as more human-like than videos of RL agents (Fig.~\ref{fig:ground-truth-human-agent}). Participants accurately detected human players above chance in both Study 1 (mean=$0.84$, std=$0.16$, n=$30$ throughout for both studies) and 2 (mean=$0.77$, std=$0.16$). On average, participants were slightly more accurate in judging humans as more human-like in Study 1 compared to Study 2 (Mann-Whitney $U=341$, $p=0.046$). We then analyzed the reported uncertainty of participants, reported on a scale of 1 (extremely certain) to 5 (extremely uncertain), Fig.~\ref{fig:ground-truth-hybrid-symbolic}. In Study 1 participants indicated a mean certainty of $2.1$ (std=$0.47$), corresponding to the level ``somewhat certain''. In Study 2, participants indicated an average certainty of $2.6$ (std=$0.84$), between ``somewhat certain'' and ``neutral''. The distributions in the two groups differed significantly (Mann–Whitney $U= 257.5$, $p=0.002$), indicating that participants were less certain of their decisions in Study 2.

Next we compared how participants judged symbolic and hybrid agents (Sec.~\ref{subsec:agent-nav}). In both studies, and in line with our prediction, human participants judge the hybrid agent to be more human-like when directly compared with symbolic agents (Fig.~\ref{fig:ground-truth-hybrid-symbolic}). At the same time, we observed substantial variability among human judges. A comparison across studies reveals an interesting pattern. In hybrid-symbolic comparisons of Study 1, on average $0.78$ of all participants judged the hybrid agent to be more human-like (std=$0.25$). This proportion is substantially higher than that of Study 2 (mean=$0.62$, std=$0.30$), with a statistically significant difference (Mann-Whitney $U =328.0$, p=$0.013$). 

This finding suggests that there are differences in the degree to which the hybrid agent is judged as more human-like between the two studies. Possible explanations could be confounds, however we eliminated potential confounds by matching characteristics of hybrid and symbolic agent videos across the two studies, making this explanation unlikely. Another possibility are effects such as anchoring or exposure bias. If this is the case, this not a confound but a finding that needs to be taken into consideration in future work. If confirmed, such effects would show judgments of human-likeness to be context-dependent. Future agents could use meta-reasoning to adapt their behavior to be more human-like given a context. Our results show that identifying behavioral features based on which human judges assess human-likeness is a key challenge that needs meticulous investigation to be better understood.

Overall, our analysis of human judgments show that, in line with expectations and on average, videos of human navigation are judged more human-like than those of artificial agents. In addition, as hypothesized in Sec.~\ref{subsec:agent-nav}, the hybrid agent is judged more human-like when directly compared with the symbolic agent (Figure \ref{fig:ground-truth-hybrid-symbolic}). These results validate our HNTT methodology, enabling us to validate our classifiers (artificial judges) against the human judgment ground truth reported in this section. Future behavioral studies can explore individual differences in human judgments of both human and artificial navigation.

\vskip -.2cm
\subsection{Evaluating the ANTT}
\label{subsec:res-main}
\vskip -.2cm

We now turn to our results of evaluating our proposed ANTT. We aim to identify classification models (Sec.~\ref{subsec:models}) that accurately capture human judgments of human-likeness. All models were trained to distinguish human from agent trajectories. Training and hyperparameter tuning was performed using 5-fold cross validation on trajectories generated by agent checkpoints and human players that were fully separate from those that generated test data. We collected human judgments (Sec.~\ref{subsec:res-groundtruth}) for test trajectories only.

\begin{figure}[t]
\vskip -0.1in
\begin{center}
\centerline{\includegraphics[width=.99\columnwidth]{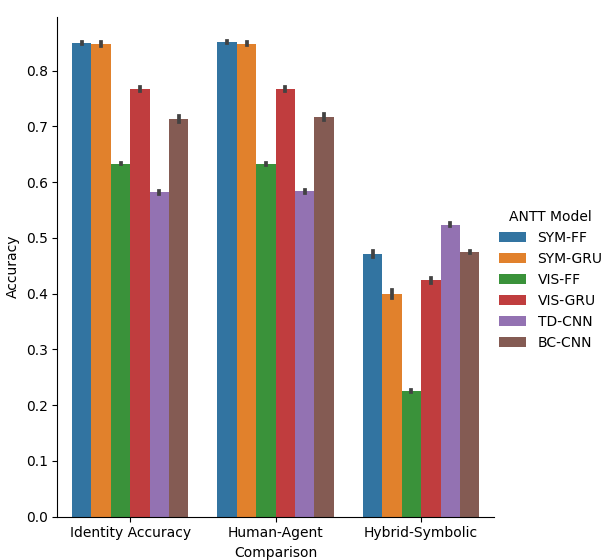}}
\vskip -0.1in
\caption{ANTT Classifier accuracy compared to ground truth identity or human judgments on held-out test data.}
\label{fig:antt-accuracy}
\end{center}
\vskip -0.3in
\end{figure}

Figure~\ref{fig:antt-accuracy} shows the accuracy of all considered classifier models (see Table~\ref{table:models-overview} for model details). First, we verify that all models generalize to the held out training data. We assess this by comparing classifier predictions (aggregated on the trajectory level by majority vote) against the true identity of the agent (i.e., human or agent) termed \textbf{identity accuracy}. We find that all models generalize to the test data, with results for the symbolic models particular robust to overfitting. Results range from an accuracy of $0.583$ for TD-CNN to $0.850$ for both symbolic models. These results also validate our trajectory level aggregation by majority voting approach, demonstrating that is achieves high accuracy when comparing to trajectory level human judgments.

Our main results compare classifier models to human judgments, split into \textbf{human-agent accuracy} and \textbf{hybrid-symbolic accuracy} (Figure~\ref{fig:antt-accuracy}). As the majority vote aggregation of human responses was equal to the true identity for all questions that compared human videos with agent videos, we again observe the same high accuracy on human-agent judgments. However, accuracy is low for hybrid-symbolic pairs across all classifiers. 

\begin{figure}[t]
\vskip -0.1in
\begin{center}
\centerline{\includegraphics[width=.99\columnwidth]{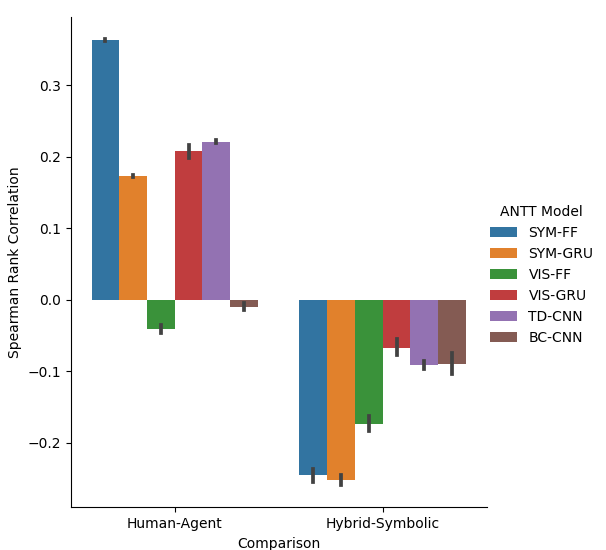}}
\vskip -0.1in
\caption{Spearman rank correlation of ANTT rankings compared to human judgments on held-out test data.}
\label{fig:antt-rank}
\end{center}
\vskip -0.3in
\end{figure}

Our results are further confirmed when assessing \textbf{human-agent} and \textbf{hybrid-symbolic rank} correlation between human judgments and model predictions (Figure~\ref{fig:antt-rank}). We compute these using Spearman's rank correlation \cite{croux2010influence}, as follows. We aggregate human ground truth judgments for human-agent (separately: symbolic-hybrid) pairs based on the percentage of human judges that labeled a human (or hybrid agent in hybrid-symbolic) trajectory as more human-like. This was compared to a ranking of model output logits. We observe medium rank agreement and low variance for the symbolic and topdown CNN models on human-agent rankings. However, rankings of all models on hybrid-symbolic agent rankings are low (negative values indicate anti-correlated rankings) and have high variance.

Our classification results demonstrate that classifier models do generalize well to unseen test trajectories. Agreement with human judgments is high when assessing the human-likeness of agents compared to human players. Highest agreement is observed for symbolic models. This relative performance of classifiers is unexpected, given that the symbolic classifiers observe a representation that is very different from that used by human judges for decision making. Based on our results, it is likely that the visual models overfit given the richer representations and higher model capacity. In addition, we find that current classifiers perform poorly when assessing the degree of human-likeness of pairs of agents. Our results motivate important follow up work to close this gap between matching human judgments on human-agent vs agent-agent pairs.

\vskip -.2cm
\section{Conclusion}
\label{sec:conclusion}
\vskip -.2cm

We introduce the automated Navigation Turing Test (ANTT), a new approach to automatically evaluating the human-likeness of navigation behavior. Our results, comparing to a human NTT, show that a high level of agreement between automated and human judgments can be achieved, opening the door to using classification approaches as scalable proxies for driving the development of more human-like agents. Importantly, we do not (and do not recommend) directly using these models to reward learning human-like policies as doing so can lead to agents that exploit static classifiers \cite{hernandez2020twenty,shi2021adversarial}.

At the same time, we show that detecting finer-grained human judgments that capture agents' improvements towards human-likeness remains an open challenge for future work. Conceptually related work by \citep{gross2017generalizing} showed enabling the discriminator to interrogate the generator improved performance over passively learning to discriminate via supervised learning alone. The ability of the discriminator to interrogate was also a core part of the original Turing Test and may be key to improving finer grained discrimination between agents.

Our work takes an important step on the dual goals of understanding human assessment of human-like behavior and towards agents capable of learning complex human-like behaviors.

\bibliography{references}
\bibliographystyle{icml2021}

\appendix

\begin{figure}[th]
\begin{center}
\includegraphics[width=.48\columnwidth]{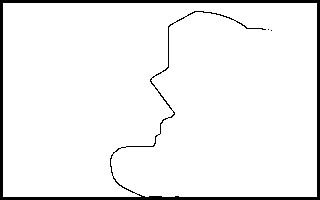}
\includegraphics[width=.48\columnwidth]{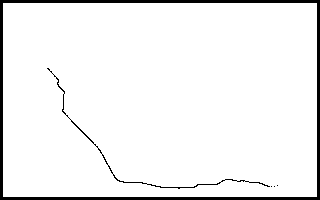}
\includegraphics[width=.48\columnwidth]{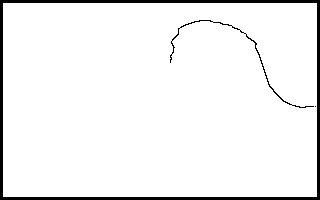}
\includegraphics[width=.48\columnwidth]{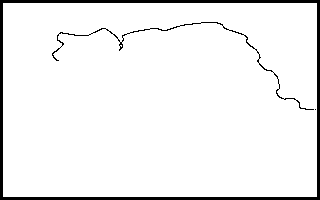}
\vskip -0.05in
\caption{Inputs for the TD-CNN model. Random samples of two agent trajectories (top) and two human trajectories (bottom). Each image represents one whole
trajectory/video, obtained by projecting the symbolic representation
(agent position) along the ”up” direction (z-coordinate).}
\label{fig:sample-td}
\end{center}
\end{figure}

\section{Appendix}

\subsection{Classifier Training Details}
\label{subsec:Classifier_appendix}

This section provides training details for our Automated Navigation Turing Test (ANTT) classifiers (described in Section 3 of the main paper).

To estimate the mean validation accuracy for hyperparameter tuning each model, we ran 5-fold cross-validation with an 80-20 split for training and validation. The training and validation sets are composed of a total of 100 episodes collected by four human players and 198 episodes of trained agents from two checkpoints. The test set is composed of the 40 videos shown in the Human Navigation Turing Test (HNTT, see Section 4 of the main paper), collected by three different human players and a different checkpoint for the trained agents (i.e., there was no overlap in players or agent checkpoints between the test and training/validation set). Human videos are selected by weighted sampling during cross-validation to account for class imbalance.

\begin{figure}[t]
\begin{center}
\includegraphics[trim=0 0 0 0,clip,width=.98\columnwidth]{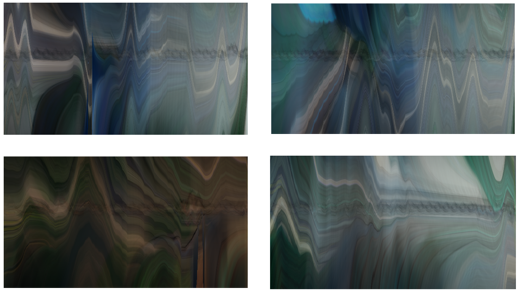}
% \vskip -0.5in
\caption{Inputs for the BC-CNN model. Random samples of two agent trajectories (top) and two human trajectories (bottom). Time is along the x-axis, each column in the image represents a single frame in the video where each colour channel has been separately averaged to compress the 2D frame to 1D in this representation. This allows us to represent an entire video the format expected by the VGG network: $[$COLOR, HEIGHT, WIDTH$]$.}
\label{fig:sample-bc}
\end{center}
\end{figure}

As discussed in the paper, our experiments consider different input formats to represent human and agent trajectories. To give readers a better understanding of the quality of the \emph{top-down} (TD) and \emph{bar-code} (BC) representations, we include additional examples in Figures \ref{fig:sample-td} and \ref{fig:sample-bc}.

For the VIS-FF, VIS-GRU, and TD-CNN models, we use a VGG network \cite{simonyan2014very} pre-trained on the Imagenet dataset \cite{deng2009imagenet}. The VGG's last layer is then replaced by a feedforward network ($1$ or $2$ layers with dropout, depending on the hyperparameters) which is trained on our dataset.

Our hyperparameter tuning focused on reducing overfitting. We considered different dropout percentages ($0\%$, $50\%$, $85\%$), hidden layer dimensions ($0$, $16$, $32$) and, for recurrent models, sequence lengths  ($5$, $10$, $20$). Training efficiency was not a priority in our hyperparameter search as training was relatively fast. Each run took about $10$ minutes on a single machine equipped with a Tesla V100 GPU and $6$ Intel Xeon E5-2690 v4 CPUs. As such, hyperparameters such as batch size, optimizer, learning rate, and number of epochs were not explored. Our final best hyperparameter settings for each model are chosen based on their mean validation accuracy. This resulted in the following hyperparameters:
\begin{itemize}
    \item SYM-FF: dropout $0\%$, hidden layer size $32$, $50$ epochs, batch size $256$, Adam optimizer with learning rate $10^{-3}$.
    \item SYM-GRU: dropout $0\%$, hidden layer size $32$, sequence length $5$, $50$ epochs, batch size $256$, Adam optimizer with learning rate $10^{-3}$.
    \item VIS-FF: dropout $50\%$, hidden layer size $32$, $10$ epochs, batch size $8$, Adam optimizer with learning rate $10^{-4}$.
    \item VIS-GRU: dropout $50\%$, hidden layer size $32$, sequence length $20$, $10$ epochs, batch size $8$, Adam optimizer with learning rate $10^{-4}$.
    \item TD-CNN: dropout $50\%$, no hidden layer, $10$ epochs, batch size $32$, SGD optimizer with learning rate $5\times10^{-3}$ and momentum $0.9$.
    \item BC-CNN: dropout $0\%$, hidden layer size $32$, $10$ epochs, batch size $8$, Adam optimizer with learning rate $10^{-4}$.
\end{itemize}

\subsection{Human Navigation Turing Test - Procedure}
\label{subsec:NTT_appendix}

\begin{figure}[t]
\begin{center}
\fbox{
\begin{minipage}{7.5 cm}
\centerline{\includegraphics[width=\columnwidth]{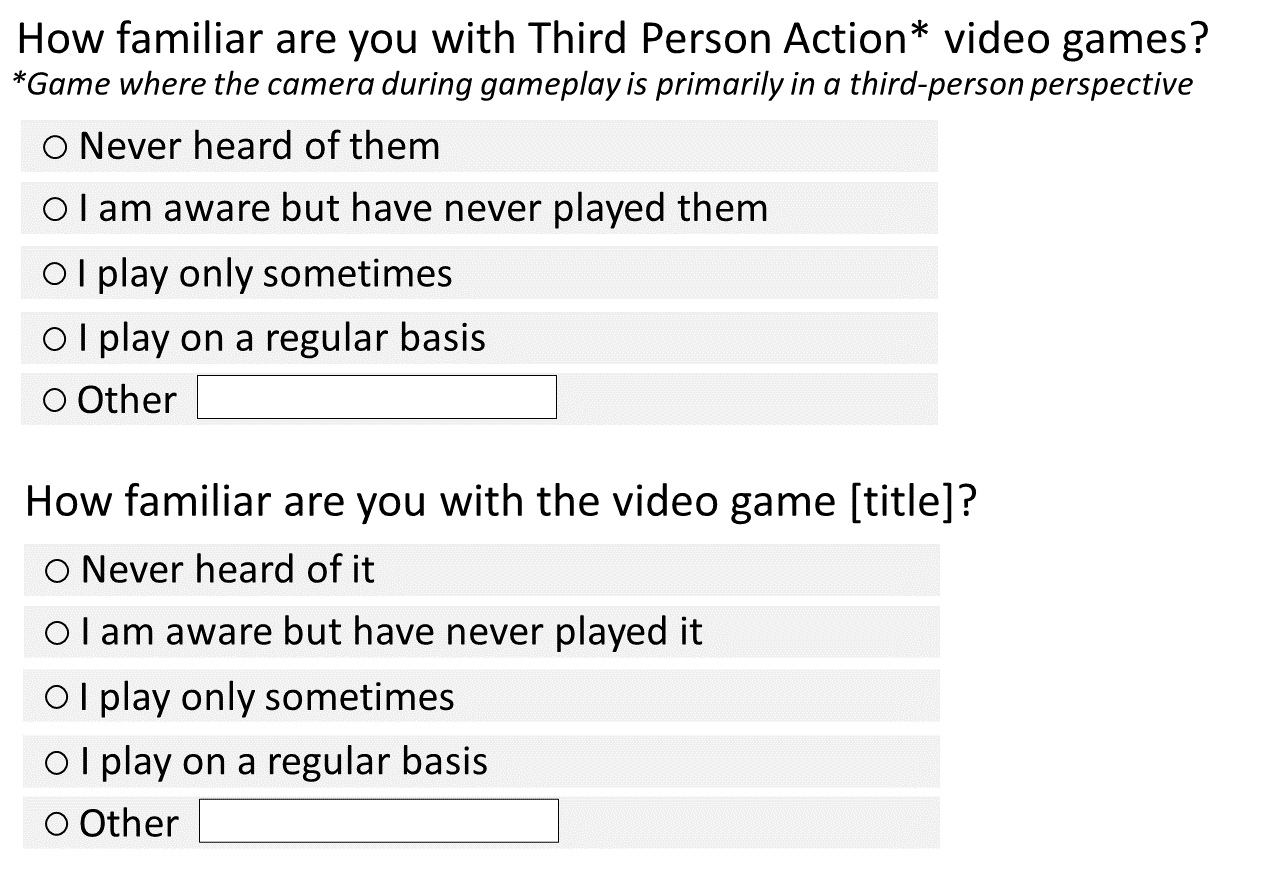}}
\end{minipage}
}
\caption{HNTT familiarity questions.}
\label{fig:study-questions}
\end{center}
\vskip -0.2in
\end{figure}

This section provides additional details about the behavioral study (Section 4.2 of the main paper) used to collect human ground truth data for our Human Navigation Turing Test (HNTT).

\textbf{Survey.} Two HNTT studies were administered as anonymous surveys structured with Introduction, Background, and Task components as follows. The Introduction included an IRB-approved consent form and was followed by a Background page, which included a short description of Third Person Action Games, the game used in this study, and the research and task descriptions. Participants were asked to rank on a 5-point Likert scale the answer to:  ``How familiar are you with Third Person Action video games?" and ``How familiar are you with the video game [title]?" (Figure \ref{fig:study-questions}).

 \textbf{Task}. The Task component of the survey consisted of ten Human Navigation Turing Test trials, in each of which participants watched two side-by-side videos (Video A and Video B), and were asked three questions about the videos. The first HNTT question, a two alternative forced choice (2AFC), was: ``which video is more likely to be human?", to which the participant could respond by choosing ``Video A is more likely to be human" or ``Video B is more likely to be human". Participants followed by giving a free-form response to the question ``why do you think this is the case? Please provide as much detail as possible". Finally they were asked to indicate on a 5-point Likert scale: ``How certain are you of your choice?". See Figure 6 in the main paper for a screenshot of the HNTT trial.

\subsection{Navigation Agent Training Details}
\label{subsec:Agent_appendix}

This section provides details on our training procedure for the reinforcement learning agents (Section 4.3 in the main paper).

\textbf{Agent architectures} Our \emph{symbolic} and \emph{hybrid} agent architectures are referenced in Figure 4 in the main paper, containing hidden layer sizes for the fully connected layers. For the convolutional layers of the hybrid model, we used the following hyperparameters:

\begin{table}[h]
\centering
\resizebox{.8\columnwidth}{!}{
\begin{tabular}{ l r }
\toprule
 \textbf{Hyperparameter}  & \textbf{Value} \\ 
\midrule\midrule
Batch size & 2048  \\ 
\midrule
Dropout rate & 0.1  \\
\midrule
Learning rate & 3e-4  \\ 
\midrule
Optimizer & Adam  \\ 
\midrule
Gamma & 0.996  \\
\midrule
Lambda & 0.95\\
\midrule
Clip range & 0.2  \\
\midrule
Gradient norm clipping coefficient & 0.5  \\ 
\midrule
Entropy coefficient & 0.0  \\ 
\midrule
Value function coefficient & 0.5 \\
\midrule
Minibatches per update & 4 \\ 
\midrule
Training epochs per update & 4 \\ 
\bottomrule
\end{tabular}
}% end resizebox
\caption{Hyperparameters for training the symbolic and the hybrid agent models using PPO \cite{schulman2017proximal}.}
\label{table:agent-hyperparam}
\end{table}

\begin{table*}[t]
\centering
\vskip -.2cm
\resizebox{1.98\columnwidth}{!}{
\begin{tabular}{lrrrrr}
\toprule
\multicolumn{1}{c}{Classifier} & \multicolumn{1}{c}{Identity Accuracy} & \multicolumn{1}{c}{Human-Agent Accuracy} & \multicolumn{1}{c}{Human-Agent Rank} & \multicolumn{1}{c}{Hybrid-Symbolic Accuracy} & \multicolumn{1}{c}{Hybrid-Symbolic Rank} \\
\midrule
SYM-FF & $0.850$ $(0.062)$ & $0.850$ $(0.062)$ & $0.364$ $(0.043)$ & $0.475$ $(0.166)$ & $-0.244$ $(0.252)$ \\
SYM-GRU & $0.850$ $(0.082)$ & $0.850$ $(0.082)$ & $0.173$ $(0.049)$ & $0.400$ $(0.200)$ & $-0.249$ $(0.210)$ \\
\midrule
VIS-FF  & $0.633$ $(0.041)$ & $0.633$ $(0.041)$ & $-0.041$ $(0.160)$ & $0.225$ $(0.050)$ & $-0.165$ $(0.286)$ \\
VIS-GRU & $0.767$ $(0.097)$ & $0.767$ $(0.097)$ & $0.220$ $(0.267)$ & $0.425$ $(0.127)$ & $-0.056$ $(0.331)$ \\
\midrule
TD-CNN & $0.583$ $(0.075)$ & $0.583$ $(0.075)$ & $0.222$ $(0.059)$ & $0.525$ $(0.094)$ & $-0.093$ $(0.149)$ \\
BC-CNN & $0.717$ $(0.145)$ & $0.717$ $(0.145)$ & $-0.009$ $(0.131)$ & $0.475$ $(0.050)$ & $-0.095$ $(0.412)$ \\
\bottomrule
\end{tabular}
}% end resizebox
\vskip -.2cm
\caption{Classifier accuracy and rank compared to human judgments on held-out test data. All results are the mean (and standard deviation) from 5 repeats of training the classifier with hyperparameter settings chosen by their average validation accuracy in 5-fold cross-validation.}
\label{table:results-classifiers}
\vskip -.1cm
\end{table*}

The output of the convolutional layers was flattened and passed through a Dense layer of size 128 with ReLU activations \cite{zeiler2013rectified}.

The size of the models' logit output is equivalent to the agents' discretized action space of size 8, which corresponds to the following valid actions: none, forward, left/right (by $30$, $45$, $90$ degrees).

We used a fixed set of hyperparameters throughout all agent training, as shown in Table~\ref{table:agent-hyperparam}. These were found to perform best on preliminary experiments.

\textbf{Training framework} Both symbolic and hybrid agents were trained using the OpenAI Baselines PPO2 implementation \cite{baselines} running on Tensorflow 2.3 \cite{tensorflow2015-whitepaper}, on top of a custom library for asynchronous data sampling.

\textbf{Training infrastructure} The symbolic model was trained on a CPU-only machine, with 64 Intel Xeon E5-2673 v4 2.3 GHz cores. The hybrid model made use of 1 GPU for training, an Nvidia Tesla K80 and 24 Intel Xeon E5-2690 v3 CPUs. The samples were collected from 60 parallel game instances, running in an Azure virtual scale set of 20 virtual machines (VMs). Each VM ran 3 separate game instances. Each simulation VM had one half of an Nvidia Tesla M60 GPU and 6 Intel Xeon E5-2690 v3 (Haswell) CPUs.

\subsection{Evaluation Details}
\label{subsec:Eval_appendix}
This section provides details for Section 5.2 in the main paper. All evaluation in this section is only on the held out test data set composed of the videos shown in the behavioral study. These videos were collected by different human players and a different checkpoint for the trained agents  than those included in the training and validation data set.

To compare our models (that are trained to classify a single trajectory as either human or agent) and the responses from the behavioral study (where participants chose which of a pair of videos was more likely to be human) we must define a method for the models to pick the most human-like video from a pair. For models that classify a single full trajectory (TD-CNN and BC-CNN) we choose the video the model predicts to have the highest likelihood of being human. For models that classify sub-sequences from a video (SYM-FF, SYM-GRU, VIS-FF and VIS-GRU) we predict the class of every non-overlapping sub-sequence, then pick the video with the highest percentage of human sub-sequence classifications (i.e., we aggregate by the robust method of majority voting). This process gives us model responses to each question in the behavioral study that can then be compared to the participant responses.
 
For questions in the behavioral study that compared one human and one agent video (questions 1 to 6 in both studies, but note that questions were presented to participants in randomized order) we calculate:

{\bf Identity Accuracy:} the accuracy of the model compared to the known origin of the video/trajectory (i.e., whether the trajectory was truly generated by a human player). 

{\bf Human-Agent Accuracy:} the accuracy of the models compared to the majority of study participants (i.e. we aggregate participant responses per question by majority vote.) 

{\bf Human-Agent Rank:} the Spearman rank correlation coefficient \cite{croux2010influence} between two lists, each with one entry per question in the behavioral study. The first is ranked by the percentage of participants that agreed with the participants' aggregated majority vote choice. The second is ranked by either the likelihood (for TD-CNN and BC-CNN) or percentage of sub-sequences classified as human (for SYM-FF, SYM-GRU, VIS-FF and VIS-GRU) for the video chosen by the model as most likely to be human.

For questions in the behavioral study that included two agents (questions 7 to 10 in both studies) we calculate {\bf Hybrid-Symbolic Accuracy} and {\bf Hybrid-Symbolic Rank} which are equivalent to the corresponding metric for the human-agent questions. For these questions, there is no equivalent metric to Identity Accuracy as both videos are from agents and so the only comparison possible is to the human ground truth data obtained through out HNTT.

For each evaluation metric that we report in Figures 9 and 10 of the main paper, the mean (and standard deviation) by averaging the value measured for each of the five trained instances of a model on the five folds of the training data with the best hyperparameters obtained by 5-fold cross validation, as detailed in Section~\ref{subsec:Classifier_appendix}. None of these models were trained on data from the test data set or other data from the same human players and agent checkpoints. For completeness, we include the raw data used to generate Figures 9 and 10 in Table~\ref{table:results-classifiers}.

\section*{Acknowledgments}
We thank Batu Aytemiz, Dave Bignell, Mikhail Jacob, Mingfei Sun, Robert Loftin, Sebastian Nowozin, and Stephanie Reikine for their feedback and valuable discussions. We are grateful to Ninja Theory and Microsoft for their support. We thank the anonymous participants of our human NTT studies for their time and effort.

All data collected throughout this study, plus the code to reproduce our analysis and ANTT are openly available at \url{https://github.com/microsoft/NTT}

\end{document}